\documentclass[11pt,a4paper]{article} 

\usepackage[hyperref]{naaclhlt2018}
\hypersetup{colorlinks=true,citecolor=darkblue, linkcolor=darkblue, urlcolor=darkblue, breaklinks=True} 

\aclfinalcopy

\usepackage{times}
\usepackage{latexsym}
\usepackage{amsmath}
\usepackage{graphicx}
\usepackage{multirow}
\usepackage{url}
\usepackage{booktabs}
\usepackage{framed}

\makeatletter
\newcommand{\@emptybiblabel}[1]{}
\makeatother

\newcommand{\fulllabel}[3]{\textsc{#1}\newline\textbf{#2}\newline\textsc{#3}}

\title{A Broad-Coverage Challenge Corpus for\\ Sentence Understanding through Inference} 

\author{
 Adina Williams$^{1}$ \\
 \texttt{\small adinawilliams@nyu.edu} \\
 \And
 Nikita Nangia$^{2}$ \\
 \texttt{\small nikitanangia@nyu.edu} \\
 \And
 Samuel R.~Bowman$^{1,2,3}$ \\
 \texttt{\small bowman@nyu.edu} \\
 \AND
 $^{1}$\normalfont Department of Linguistics\\ New York University\And
 $^{2}$\normalfont Center for Data Science\\ New York University\And
 $^{3}$\normalfont Department of Computer Science\\ New York University
}

\date{}

\begin{document}
\maketitle
\begin{abstract}
This paper introduces the Multi-Genre Natural Language Inference (MultiNLI) corpus, a dataset designed for use in the development and evaluation of machine learning models for sentence understanding. At 433k examples,  this resource is one of the largest corpora available for natural language inference (a.k.a. \textit{recognizing textual entailment}), improving upon available resources in both its coverage and difficulty. MultiNLI accomplishes this by offering data from ten distinct genres of written and spoken English, making it possible to evaluate systems on nearly the full complexity of the language, while supplying an explicit setting for evaluating cross-genre domain adaptation. In addition, an evaluation using existing machine learning models designed for the Stanford NLI corpus shows that it represents a substantially more difficult task than does that corpus, despite the two showing similar levels of inter-annotator agreement.
\end{abstract}

\section{Introduction}

\begin{table*}[t!]\label{examples}
  \centering\small
  \begin{tabular}{p{7cm}p{2.1cm}p{5.35cm}}
  \toprule
Met my first girlfriend that way. & \fulllabel{Face-to-Face}{contradiction}{c c n c} & I didn't meet my first girlfriend until later.\\
\rule{0pt}{3ex}8 million in relief in the form of emergency housing. & \fulllabel{Government}{neutral}{n n n n} & The 8 million dollars for emergency housing was still not enough to solve the problem.\\
\rule{0pt}{3ex}Now, as children tend their gardens, they have a new appreciation of their relationship to the land, their cultural heritage, and their community. & \fulllabel{Letters}{neutral}{n n n n} & All of the children love working in their gardens.\\
\rule{0pt}{3ex}At 8:34, the Boston Center controller received a third transmission from American 11 & \fulllabel{9/11}{entailment}{e e e e} & The Boston Center controller got a third transmission from American 11. \\
\rule{0pt}{3ex}I am a lacto-vegetarian. & \fulllabel{Slate}{neutral}{n n e n} & I enjoy eating cheese too much to abstain from dairy.\\
\rule{0pt}{3ex}someone else noticed it and i said well i guess that's true and it was somewhat melodious in other words it wasn't just you know it was really funny & \fulllabel{Telephone}{contradiction}{c c c c} & No one noticed and it wasn't funny at all. \\
    \bottomrule
  \end{tabular}
  \caption{Randomly chosen examples from the development set of our new corpus, shown with their genre labels, their selected gold labels, and the validation labels (abbreviated \textsc{e, n, c}) assigned by individual annotators.}
\end{table*}

Many of the most actively studied problems in NLP, including question answering, translation, and dialog, depend in large part on natural language understanding (NLU) for success. While there has been a great deal of work that uses representation learning techniques to pursue progress on these applied NLU problems directly, in order for a representation learning model to fully succeed at one of these problems, it must simultaneously succeed both at NLU, and at one or more additional hard machine learning problems like structured prediction or memory access. This makes it difficult to accurately judge the degree to which current models extract reasonable representations of language meaning in these settings.

The task of natural language inference (NLI) is well positioned to serve as a benchmark task for research on NLU. In this task, also known as \textit{recognizing textual entailment} \cite{Fyodorov-etal:2000,Condoravdi-etAl:2003,BosMar:2005,dagan2006pascal,maccartney2009extended}, a model is presented with a pair of sentences---like one of those in Figure~\ref{examples}---and asked to judge the relationship between their meanings by picking a label from a small set: typically \textsc{entailment}, \textsc{neutral}, and \textsc{contradiction}. Succeeding at NLI does not require a system to solve any difficult machine learning problems except, crucially, that of extracting an effective and thorough representations for the meanings of sentences (i.e., their lexical and compositional semantics). In particular, a model must handle phenomena like lexical entailment, quantification, coreference, tense, belief, modality, and lexical and syntactic ambiguity.  


As the only large human-annotated corpus for NLI currently available, the Stanford NLI Corpus \citep[SNLI;][]{snli:emnlp2015} has enabled a good deal of progress on NLU, serving as a major benchmark for machine learning work on sentence understanding and spurring work on core representation learning techniques for NLU, such as attention \cite{wang2015learning,parikh2016},  memory \cite{munkhdalai2016neural}, and the use of parse structure \cite{mou2015recognizing,bowman-EtAl:2016:P16-1,chen2017esim}. 
 However, SNLI falls short of providing a sufficient testing ground for machine learning models in two ways. First, the sentences in SNLI are derived from only a single text genre---image captions---and are thus limited to descriptions of concrete visual scenes, rendering the hypothesis sentences used to describe these scenes short and simple, and rendering many important phenomena---like temporal reasoning (e.g., \textit{yesterday}), belief (e.g., \textit{know}), and modality (e.g., \textit{should})---rare enough to be irrelevant to task performance. Second, because of these issues, SNLI is not sufficiently demanding to serve as an effective benchmark for NLU, with the best current model performance falling within a few percentage points of human accuracy and limited room left for fine-grained comparisons between strong models.

This paper introduces a new challenge dataset, the Multi-Genre NLI Corpus (MultiNLI), whose chief purpose is to remedy these limitations by making it possible to run large-scale NLI evaluations that capture more of the complexity of modern English. While its size (433k pairs) and mode of collection are modeled closely on SNLI, unlike that corpus, MultiNLI represents both written and spoken speech in a wide range of styles, degrees of formality, and topics.

Our chief motivation in creating this corpus is to provide a benchmark for ambitious machine learning research on the core problems of NLU, but we are additionally interested in constructing a corpus that facilitates work on domain adaptation and cross-domain transfer learning. In many application areas outside NLU, artificial neural network techniques have made it possible to train general-purpose feature extractors that, with no or minimal retraining, can extract useful features for a variety of styles of data \citep{krizhevsky2012imagenet,zeiler2013visualizing,donahue2014decaf}. However, attempts to bring this kind of general purpose representation learning to NLU have seen only very limited success \citep[see, for example,][]{mou2016transferable}. Nearly all successful applications of representation learning to NLU have involved models that are trained on data that closely resembles the target evaluation data, both in task and style. This fact limits the usefulness of these tools for problems involving styles of language not represented in large annotated training sets. 

With this in mind, we construct MultiNLI so as to make it possible to explicitly evaluate models both on the quality of their sentence representations within the training domain and on their ability to derive reasonable representations in unfamiliar domains. The corpus is derived from ten different genres of written and spoken English, which are collectively meant to approximate the full diversity of ways in which modern standard American English is used.  All of the genres appear in the test and development sets, but only five are included in the training set. Models thus can be evaluated on both the \textit{matched} test examples, which are derived from the same sources as those in the training set, and on the \textit{mismatched} examples, which do not closely resemble any of those seen at training time.  


\section{The Corpus}

\subsection{Data Collection}

The data collection methodology for MultiNLI is similar to that of SNLI: We create each sentence pair by selecting a premise sentence from a preexisting text source and asking a human annotator to compose a novel sentence to pair with it as a hypothesis. This section discusses the sources of our premise sentences, our collection method for hypotheses, and our validation (relabeling) strategy.

\paragraph{Premise Text Sources}

The MultiNLI premise sentences are derived from ten sources of freely available text which are meant to be maximally diverse and roughly represent the full range of American English.  We selected nine sources from the second release of the Open American National Corpus \citep[OANC; ][downloaded 12/2016\footnote{ \url{http://www.anc.org/}}]{fillmore1998,macleod2000,ide2001, ide2006openANC}, balancing the volume of source text roughly evenly across genres, and avoiding genres with content that would be too difficult for untrained annotators. 
 
 \begin{figure}[t] 
 	\begin{framed}\small
 		This task will involve reading a line from a non-fiction article and writing three sentences that relate to it. The line will describe a situation or event. Using only this description and what you know about the world:
 		\begin{itemize}
 			\item Write one sentence that is definitely correct about the situation or event in the line. 
 			\item Write one sentence that might be correct about the situation or event in the line. 
 			\item Write one sentence that is definitely incorrect about the situation or event in the line. 
 		\end{itemize}
 	\end{framed}
 	\caption{\label{hybridprompt}The main text of a prompt (truncated) that was presented to our annotators. This version is used for the written non-fiction genres.} 
 \end{figure}
 
OANC data constitutes the following nine genres: {transcriptions from \href{http://newsouthvoices.uncc.edu/}{the Charlotte Narrative and Conversation Collection} of two-sided, in-person conversations that took place in the early 2000s (\textsc{Face-to-face}); reports, speeches, letters, and press releases from public domain government websites (\textsc{Government}); letters from \href{https://liberalarts.iupui.edu/icic/research/corpus_of_philanthropic_fundraising_discourse}{the Indiana Center for Intercultural Communication of Philanthropic Fundraising Discourse} written in the late 1990s--early 2000s  (\textsc{Letters}); the public report from \href{https://9-11commission.gov/}{the National Commission on Terrorist \-Attacks Upon the United States} released on July 22, 2004\footnote{\url{https://9-11commission.gov/}} (\textsc{9/11}); five non-fiction works on the textile industry and child development published by the Oxford University Press (\textsc{OUP}); popular culture articles from the archives of Slate Magazine  (\textsc{Slate}) written between 1996--2000; transcriptions from University of Pennsylvania's \href{https://catalog.ldc.upenn.edu/LDC97S62}{Linguistic Data Consortium Switchboard corpus} of two-sided, telephone conversations that took place in 1990 or 1991 (\textsc{Telephone}); travel guides published by Berlitz Publishing in the early 2000s (\textsc{Travel}); and short posts about linguistics for non-specialists from the \href{http://www.verbatimmag.com/}{Verbatim} archives written between 1990 and 1996 (\textsc{Verbatim}). 

For our tenth genre, \textsc{Fiction}, we compile several freely available works of contemporary fiction written between 1912 and 2010, spanning genres including mystery, humor, western, science fiction, and fantasy by authors  Isaac Asimov, Agatha Christie, Ben Essex (Elliott Gesswell), Nick Name (Piotr Kowalczyk), Andre Norton, Lester del Ray, and \href{http://mikeshea.net}{Mike Shea}. 

We construct premise sentences from these ten source texts with minimal preprocessing; unique the sentences within genres, exclude very short sentences (under eight characters), and manually remove certain types of non-narrative writing, such as mathematical formulae, bibliographic references, and lists. 

Although SNLI is collected in largely the same way as MultiNLI, and is also permissively licensed, we do not include SNLI in the MultiNLI corpus distribution. SNLI can be appended and treated as an unusually large additional \textsc{captions} genre, built on image captions from the Flickr30k corpus \cite{hodoshimage}.

\paragraph{Hypothesis Collection}

To collect a sentence pair, we present a crowdworker with a sentence from a source text and ask them to compose three novel sentences (the hypotheses): one which is necessarily true or appropriate whenever the premise is true (paired with the premise and labeled \textsc{entailment}), one which is necessarily false or inappropriate whenever the premise is true (\textsc{contradiction}), and one where neither condition applies (\textsc{neutral}). This method of data collection ensures that the three classes will be represented equally in the raw corpus.

The prompts that surround each premise sentence during hypothesis collection are slightly tailored to fit the genre of that premise sentence. We pilot these prompts prior to data collection to ensure that the instructions are clear and that they yield hypothesis sentences that fit the intended meanings of the three classes. There are five unique prompts in total: one for written non-fiction genres (\textsc{Slate, OUP, Government, Verbatim, Travel}; Figure~\ref{hybridprompt}), one for spoken genres (\textsc{Telephone, Face-to-Face}), one for each of the less formal written genres (\textsc{Fiction, Letters}), and a specialized one for \textsc{9/11}, tailored to fit its potentially emotional content. Each prompt is accompanied by example premises and hypothesis that are specific to each genre.

Below the instructions, we present three text fields---one for each label---followed by a field for reporting issues, and a link to the frequently asked questions (FAQ) page. We provide one FAQ page per prompt. FAQs are modeled on their SNLI counterparts (supplied by the authors of that work) and include additional curated examples, answers to genre-specific questions arising from our pilot phase, and information about logistical concerns like payment. 

For both hypothesis collection and validation, we present prompts to annotators using \href{gethybrid.io}{Hybrid} (\url{gethybrid.io}), a crowdsoucring platform similar to the Amazon Mechanical Turk platform used for SNLI. We used this platform to hire an organized group of workers. 387 annotators contributed through this group, and at no point was any identifying information about them, including demographic information, available to the authors. 

\begin{table}[t]
	\begin{centering}
		\small
		\begin{tabular}{lrr} 
			\toprule
			\bf Statistic & \bf SNLI & \bf MultiNLI\\
			\midrule
			Pairs w/ unanimous gold label & 58.3\% & 58.2\%\\
			\midrule
			Individual label $=$ gold label & 89.0\% & 88.7\%\\
			Individual label $=$ author's label & 85.8\% & 85.2\% \\
			\midrule
			Gold label $=$ author's label & 91.2\% & 92.6\% \\
			Gold label $\ne$ author's label & 6.8\% & 5.6\% \\
			No gold label (no 3 labels match) & 2.0\% & 1.8\%\\
			\bottomrule
		\end{tabular}
		\caption{\label{tab:validation-stats}Key validation statistics for SNLI \citep[copied from][]{snli:emnlp2015} and MultiNLI.}
	\end{centering}
\end{table}

\begin{table*}[htbp]
	\centering\small 
	\begin{tabular}{lrrrcrrrrr}
		\toprule
		\bf  & \multicolumn{3}{c}{\bf \#Examples} & \bf \#Wds. & \multicolumn{2}{c}{\bf `S' parses} &  & \multicolumn{2}{c}{\bf Model Acc.} \\
		\bf Genre & \bf Train& \bf Dev. & \bf Test & \bf Prem. & \bf Prem. & \bf Hyp. &  \bf Agrmt.& \bf ESIM & \bf CBOW \\
		\midrule
		\it SNLI & \it 550,152 & \it 10,000 & \it 10,000 & \it 14.1 & \it 74\%& \it 88\% & \it 89.0\% & \it 86.7\% & \it 80.6 \% \\
		\midrule
		\sc Fiction & 77,348 & 2,000 & 2,000 & 14.4 & 94\%&97\% & 89.4\%& 73.0\% &\ 67.5\% \\
		\sc Government & 77,350 & 2,000 & 2,000 & 24.4 &90\%&97\%& 87.4\%& 74.8\% &\ 67.5\% \\
		\sc Slate & 77,306 & 2,000 & 2,000 & 21.4 &94\%&98\%& 87.1\%& 67.9\%  &\ 60.6\%\\
		\sc Telephone & 83,348 & 2,000 & 2,000 & 25.9 &71\%&97\%& 88.3\%& 72.2\%  &\ 63.7\%\\
		\sc Travel & 77,350 & 2,000 & 2,000 & 24.9 &97\%&98\%& 89.9\%& 73.7\%  &\ 64.6\%\\
		\midrule
		\sc 9/11 & 0 & 2,000 & 2,000 & 20.6 &98\%&99\%& 90.1\%& 71.9\%  &\ 63.2\%\\
		\sc Face-to-face & 0 & 2,000 & 2,000 & 18.1 &91\%&96\%& 89.5\%& 71.2\%  &\ 66.3\%\\
		\sc Letters & 0 & 2,000 & 2,000 & 20.0 &95\%&98\%& 90.1\%& 74.7\%  &\ 68.3\%\\
		\sc OUP & 0 & 2,000 & 2,000 & 25.7 &96\%&98\%& 88.1\%& 71.7\%  &\ 62.8\%\\
		\sc Verbatim & 0 & 2,000 & 2,000 & 28.3 &93\%&97\%& 87.3\%& 71.9\%  &\ 62.7\%\\
		\midrule
		\bf MultiNLI Overall & \textbf{392,702} & \textbf{20,000} & \textbf{20,000} & \bf 22.3 & \bf 91\%&\bf 98\% & \bf 88.7\% & \bf 72.2\% &\ \bf 64.7\%\\
		\bottomrule
	\end{tabular}
	\caption{\label{tab:stats}Key statistics for the corpus by genre. The first five genres represent the \textit{matched} section of the development and test sets, and the remaining five represent the \textit{mismatched} section. The first three statistics provide the number of examples in each genre. \textit{\#Wds. Prem.}~is the mean token count among premise sentences. \textit{`S' parses} is the percentage of sentences for which the Stanford Parser produced a parse rooted with an `S' (sentence) node. \textit{Agrmt.}~is the percent of individual labels that match the gold label in validated examples. \textit{Model Acc.}~gives the test accuracy for ESIM and CBOW models (trained on either SNLI or MultiNLI), as described in Section~\ref{sec:baselines}.}
\end{table*}

\paragraph{Validation}
We perform an additional round of annotation on test and development examples to ensure accurate labelling. The validation phase follows the same procedure used for SICK \cite{marelli2014sick} and SNLI: Workers are presented with pairs of sentences and asked to supply a single label (\textsc{entailment}, \textsc{contradiction}, \textsc{neutral}) for the pair.  Each pair is relabeled by four workers, yielding a total of five labels per example. Validation instructions are tailored by genre, based on the main data collection prompt (Figure \ref{hybridprompt}); a single FAQ, modeled after the validation FAQ from SNLI, is provided for reference. In order to encourage thoughtful labeling, we manually label one percent of the validation examples and offer a \$1 bonus each time a worker selects a label that matches ours.

For each validated sentence pair, we assign a \textit{gold label} representing a majority vote between the initial label assigned to the pair by the original annotator, and the four additional labels assigned by validation annotators. A small number of examples did not receive a three-vote consensus on any one label. These examples are included in the distributed corpus, but are marked with `\texttt{-}' in the gold label field, and should not be used in standard evaluations. Table~\ref{tab:validation-stats} shows summary statistics capturing the results of validation, alongside corresponding figures for SNLI. These statistics indicate that the labels included in MultiNLI are about as reliable as those included in SNLI, despite MultiNLI's more diverse text contents. 

\subsection{The Resulting Corpus}

Table \ref{examples} shows randomly chosen development set examples from the collected corpus. Hypotheses tend to be fluent and correctly spelled, though not all are complete sentences. Punctuation is often omitted. Hypotheses can rely heavily on knowledge about the world, and often don't correspond closely with their premises in syntactic structure. 

Unlabeled test data is available on Kaggle for both  \href{https://www.kaggle.com/c/multinli-matched-open-evaluation}{matched} and \href{https://www.kaggle.com/c/multinli-mismatched-open-evaluation}{mismatched} sets as competitions that will be open indefinitely; Evaluations on a subset of the test set have previously been conducted with different leaderboards through the \href{https://repeval2017.github.io/shared/}{RepEval 2017 Workshop} \citep{nangia2017}. 



The corpus is available in two formats---tab separated text and JSON Lines (\texttt{jsonl}), following SNLI. For each example, premise and hypothesis strings, unique identifiers for the pair and prompt, and the following additional fields are specified:
\begin{itemize}
\item \texttt{gold\_label}: label used for classification. In examples rejected during the validation process, the value of this field will be `\texttt{-}'.
\item \texttt{sentence\{1,2\}\_parse}: Each sentence as parsed by the Stanford PCFG Parser 3.5.2 \cite{klein2003accurate}.
\item \texttt{sentence\{1,2\}\_binary\_parse}: parses in unlabeled binary-branching format.
\item \texttt{label[1]}: The label assigned during the creation of the sentence pair. In rare cases this may be different from \texttt{gold\_label}, if a consensus of annotators chose a different label during the validation phase.
\item \texttt{label[2...5]}: The four labels assigned during validation by individual annotators to each development and test example. These fields will be empty for training examples.
\end{itemize}

The current version of the corpus is freely available at \url{nyu.edu/projects/bowman/multinli/} for typical machine learning uses, and may be modified and redistributed. The majority of the corpus is released under the OANC's license, which allows all content to be freely used, modified, and shared under permissive terms. The data in the \textsc{Fiction} section falls under several permissive licenses; \textit{Seven Swords} is available under a Creative Commons Share-Alike 3.0 Unported License, and with the explicit permission of the author, \textit{Living History} and \textit{Password Incorrect} are available under Creative Commons Attribution 3.0 Unported Licenses; the remaining works of fiction are in the public domain in the United States (but may be licensed differently elsewhere). 

\paragraph{Partition} The distributed corpus comes with an explicit train/test/development split. The test and development sets contain 2,000 randomly selected examples each from each of the genres, resulting in a total of 20,000 examples per set. No premise sentence occurs in more than one set.  

\paragraph{Statistics} Table~\ref{tab:stats} shows some additional statistics. Premise sentences in MultiNLI tend to be longer (max 401 words, mean 22.3 words) than their hypotheses (max 70 words), 
and much longer, on average, than premises in SNLI (mean 14.1 words); premises in MultiNLI also tend to be parsed as complete sentences at a much higher rate on average (91\%) than their SNLI counterparts (74\%). We observe that the two spoken genres differ in this---with \textsc{Face-to-face} showing more complete sentences (91\%) than \textsc{Telephone} (71\%)---and speculate that the lack of visual feedback in a telephone setting may result in a high incidence of interrupted or otherwise incomplete sentences. 

Hypothesis sentences in MultiNLI generally cannot be derived from their premise sentences using only trivial editing strategies. While $2.5$\% of the hypotheses in SNLI differ from their premises by deletion, only $0.9$\% of those in MultiNLI (170 examples total) are constructed in this way. Similarly, in SNLI, $1.6$\% of hypotheses differ from their premises by addition, substitution, or shuffling a single word, while in MultiNLI this only happens in $1.2$\% of examples. The percentage of hypothesis-premise pairs with high token overlap ($>$37\%) was comparable between MultiNLI (30\% of pairs) and SNLI (29\%). These statistics suggest that MultiNLI's annotations are comparable in quality to those of SNLI.

\section{Baselines}\label{sec:baselines}
\begin{table}[t!] 
	\centering\small
	\begin{tabular}{llcccc}
		\toprule
		&  & & \multicolumn{2}{c}{\bf MNLI} \\ 
		\bf Train & \bf Model & \bf SNLI & \bf Match. & \bf Mis. \\ 
		\midrule
		& Most freq.  & 34.3 &  36.5 & 35.6\\ \midrule
		\multirow{3}{*}{SNLI} & CBOW & 80.6 & - & - \\ & BiLSTM &  81.5 & - & - \\ & ESIM & \bf 86.7 & - & - \\ \midrule
		\multirow{3}{*}{MNLI} & CBOW & 51.5 & 64.8 & 64.5 \\ & BiLSTM & 50.8 & 66.9 & 66.9 \\ & ESIM & 60.7 & \bf 72.3 & \bf 72.1 \\ \midrule
		\multirow{3}{*}{\parbox{1.5cm}{MNLI+ SNLI}} & CBOW & 74.7 & 65.2 & 64.6 \\ & BiLSTM & 74.0 & 67.5 & 67.1 \\ & ESIM & 79.7 & \bf 72.4 & \bf 71.9 \\
		\bottomrule
	\end{tabular}
	\caption{\label{tab:results}Test set accuracies (\%) for all models; \textit{Match.} represents test set performance on the MultiNLI genres that are also represented in the training set, \textit{Mis.} represents test set performance on the remaining ones; \textit{Most freq.} is a trivial `most frequent class' baseline.}
\end{table}

To test the difficulty of the corpus, we experiment with three neural network models. The first is a simple continuous bag of words (CBOW) model in which each sentence is represented as the sum of the embedding representations of its words. The second computes representations by averaging the states of a bidirectional LSTM RNN \citep[BiLSTM;][]{hochreiter1997long} over words. For the third, we implement and evaluate \citeauthor{chen2017esim}'s Enhanced Sequential Inference Model (ESIM), which is roughly tied for the state of the art on SNLI at the time of writing. We use the base ESIM without ensembling with a TreeLSTM (as in the `HIM' runs in that work). 

The first two models produce separate vector representations for each sentence and compute label predictions for pairs of representations. To do this, they concatenate the representations for premise and hypothesis, their difference, and their element-wise product, following \citet{mou2015recognizing}, and pass the result to a single $\tanh$ layer followed by a three-way softmax classifier. 

All models are initialized with 300D reference GloVe vectors \citep[840B token version;][]{pennington2014glove}. Out-of-vocabulary (OOV) words are initialized randomly and word embeddings are fine-tuned during training. The models use 300D hidden states, as in most prior work on SNLI. We use Dropout \citep{srivastava2014dropout} for regularization. For ESIM, we use a dropout rate of 0.5, following the paper. For CBOW and BiLSTM models, we tune Dropout on the SNLI dev.~set and find that a drop rate of 0.1 works well. We use the Adam \citep{kingma2014adam} optimizer with default parameters. 

We train models on SNLI, MultiNLI, and a mixture; Table~\ref{tab:results} shows the results. In the mixed setting, we use the full MultiNLI training set and randomly select 15\% of the SNLI training set at each epoch, ensuring that each available genre is seen during training with roughly equal frequency. 

We also train a separate CBOW model on each individual genre to establish the degree to which simple models already allow for effective transfer across genres, using a dropout rate of 0.2. When training on SNLI, a single random sample of 15\% of the original training set is used. For each genre represented in the training set, the model that performs best on it was trained on that genre; a model trained only on SNLI performs worse on every genre than comparable models trained on any genre from MultiNLI.

Models trained on a single genre from MultiNLI perform well on similar genres; for example, the model trained on \textsc{Telephone} attains the best accuracy (63\%) on \textsc{Face-to-Face}, which was nearly one point better than it received on itself. \textsc{Slate} seems to be a difficult and relatively unusual genre and performance on it is relatively poor in this setting; when averaging over runs trained on SNLI and all genres in the matched section of the training set, average performance on \textsc{Slate} was only 57.5\%. Sentences in \textsc{Slate} cover a wide range of topics and phenomena, making it hard to do well on, but also forcing models trained on it be broadly capable; the model trained on \textsc{Slate} achieves the highest accuracy of any model on \textsc{9/11} (55.6\%) and \textsc{Verbatim} (57.2\%), and relatively high accuracy on \textsc{Travel} (57.4\%) and \textsc{Government} (58.3\%). We also observe that our models perform similarly on both the matched and mismatched test sets of MultiNLI. 
We expect genre mismatch issues to become more conspicuous as models are developed that can better fit MultiNLI's training genres.

\section{Discussion and Analysis}


\subsection{Data Collection}
In data collection for NLI, different annotator decisions about the coreference between entities and events across the two sentences in a pair can lead to very different assignments of pairs to labels \cite{de2008finding,marelli2014semeval,snli:emnlp2015}. Drawing an example from Bowman et al., the pair \textit{``a boat sank in the Pacific Ocean''} and \textit{``a boat sank in the Atlantic Ocean''} can be labeled either \textsc{contradiction} or \textsc{neutral} depending on (among other things) whether the two mentions of boats are assumed to refer to the same entity in the world. This uncertainty can present a serious problem for inter-annotator agreement, since it is not clear that it is possible to define an explicit set of rules around coreference that would be easily intelligible to an untrained annotator (or any non-expert). 

Bowman et al.~attempt to avoid this problem by using an annotation prompt that is highly dependent on the concreteness of image descriptions; but, as we engage with the much more abstract writing that is found in, for example, government documents, there is no reason to assume \textit{a priori} that any similar prompt and annotation strategy can work. We are surprised to find that this is not a major issue.  Through a relatively straightforward trial-and-error piloting phase, followed by discussion with our annotators, we manage to design prompts for abstract genres that yield high inter-annotator agreement scores nearly identical to those of SNLI (see Table \ref{tab:validation-stats}). These high scores suggest that our annotators agreed on a single task definition, and were able to apply it consistently across genres.

\subsection{Overall Difficulty}\label{sec:difficulty}

\begin{table*}[t]
	\centering\small 
	\begin{tabular}{lrrrlrrrr}
		\toprule
		& \multicolumn{3}{c}{\bf Dev. Freq.} & \multicolumn{2}{c}{\bf Most Frequent Label} & \multicolumn{3}{c}{\bf Model Acc.}  \\
		\bf Tag & \bf  SNLI & \bf MultiNLI & \bf Diff. & \bf Label & \bf \% & \bf CBOW & \bf BiLSTM  & \bf ESIM \\
		\midrule
		Entire Corpus & 100 & 100 & 0 &  entailment & \it $\sim$35 & \it $\sim$65 & \it $\sim$67 & \it $\sim$72 \\
		\midrule
		Pronouns \small{(PTB)}  & 34& 68 & 34 & entailment & 34 & 66 & 68  & 73 \\
		Quantifiers   & 33 & 63 & 30 &contradiction & 36 & 66 & 68 & 73 \\
		Modals \small{(PTB)}  & $<$1 & 28 & 28 & entailment & 35 & 65 & 67 & 72  \\
		Negation \small{(PTB)}  & 5  & 31 & 26 & contradiction& \bf 48 & 67 & 70 & 75 \\
		WH terms \small{(PTB)} & 5& 30 & 25 & entailment & 35 & 64 & 65 & 72  \\
		Belief Verbs & $<$1 & 19 & 18 & entailment & 34 & 64 & 67 & 71 \\
		Time Terms  & 19  & 36 & 17 &neutral & 35 &64 & 66 & 71 \\
		Discourse Mark. & $<$1  & 14 & 14 & neutral & 34 & 62 & 64 & 70 \\
		Presup. Triggers & 8  & 22 & 14 & neutral & 34 & 65 & 67 & 73 \\
		Compr./Supr.\small{(PTB)}   & 3  & 17 & 14 & neutral & 39 & 61 & 63 & 69 \\
		Conditionals & 4 & 15 &11 & neutral & 35 & 65 & 68 & 73  \\
		Tense Match (PTB) & 62  & 69 & 7 & entailment & 37 & 67 & 68 & 73 \\
		Interjections  \small{(PTB)}  &  $<$1 & 5 & 5 & entailment & 36 & 67 & 70 & 75 \\
		$>$20 words  & $<$1 & 5 & 5 & entailment & \bf 42 &65 & 67 & 76 \\
		\bottomrule
	\end{tabular}
	\caption{\label{tab:autotag}Dev. Freq. is the percentage of dev.~set examples that include each phenomenon, ordered by greatest difference in frequency of occurrence (Diff.) between MultiNLI and SNLI. Most Frequent Label specifies which label is the most frequent for each tag in the MultiNLI dev.~set, and \%  is its incidence. Model Acc. is the dev.~set accuracy (\%) by annotation tag for each baseline model (trained on MultiNLI only). (PTB) marks a tag as derived from Penn Treebank-style parser output tags \citep[][]{ptb}. 
	}
\end{table*}

As expected, both the increase in the diversity of linguistic phenomena in MultiNLI and its longer average sentence length conspire to make MultiNLI dramatically more difficult than SNLI. Our three baseline models perform better on SNLI than MultiNLI by about 15\% when trained on the respective datasets. All three models achieve accuracy above 80\% on the SNLI test set when trained only on SNLI. However, when trained on MultiNLI, only ESIM surpasses 70\% accuracy on MultiNLI's test sets. When we train models on MultiNLI and downsampled SNLI, we see an expected significant improvement on SNLI, but no significant change in performance on the MultiNLI test sets, suggesting including SNLI in training doesn't drive substantial improvement. These results attest to MultiNLI's difficulty, and with its relatively high inter-annotator agreement, suggest that it presents a problem with substantial headroom for future work. 

\subsection{Analysis by Linguistic Phenomenon} 

To better understand the types of language understanding skills that MultiNLI tests, we analyze the collected corpus using a set of annotation tags chosen to reflect linguistic phenomena which are known to be potentially difficult. We use two methods to assign tags to sentences. First, we use the Penn Treebank \citep[PTB;][]{ptb} part-of-speech tag set (via the included Stanford Parser parses) to automatically isolate sentences containing a range of easily-identified phenomena like comparatives. Second, we isolate sentences that contain hand-chosen key words indicative of additional interesting phenomena.

The hand-chosen tag set covers the following phenomena: \textsc{Quantifiers} contains single words with quantificational force \citep[see, for example,][e.g., \textit{many, all, few, some}]{heimkratzer,Szabolcsi:2010}; 
 \textsc{Belief Verbs} contains sentence-embedding verbs denoting mental states (e.g., \textit{know, believe, think}), including irregular past tense forms; 
 \textsc{Time terms} contains single words with abstract temporal interpretation, (e.g., \textit{then, today}) and month names and days of the week; \textsc{Discourse markers} contains words that facilitate discourse coherence (e.g., \textit{yet, however, but, thus, despite}); \textsc{Presupposition Triggers} contains words with lexical presuppositions \citep[][e.g., \textit{again, too, anymore\footnote{Because their high frequency in the corpus, extremely common triggers like \textit{the} were excluded from this tag.}}]{stalnaker,schlenker};  \textsc{Conditionals} contains the word \textit{if}. Table \ref{tab:autotag} presents the frequency of the tags in SNLI and MultiNLI, and model accuracy on MultiNLI (trained only on MultiNLI). 

The distributions of labels within each tagged subset of the corpus roughly mirrors the balanced overall distribution. The most frequent class overall (in this case, \textsc{entailment}) occurs with a frequency of roughly one third (see Table \ref{tab:results}) in most. Only two annotation tags differ from the baseline percentage of the most frequent class in the corpus by at least 5\%: sentences containing negation, and sentences exceeding 20 words.  Sentences that contain negation are slightly more likely than average to be labeled \textsc{contradiction}, reflecting a similar finding in SNLI, while long sentences are slightly more likely to be labeled \textsc{entailment}. 

None of the baseline models perform substantially better on any tagged set than they do on the corpus overall, with average model accuracies on sentences containing specific tags falling within about 3 points of overall averages. Using baseline model test accuracy overall as a metric (see Table \ref{tab:results}), our baseline models had the most trouble on sentences containing comparatives or superlatives (losing 3-4 points each). Despite the fact that 17\% of sentence pairs in the corpus contained at least one instance of comparative or superlative, our baseline models don't utilize the information present in these sentences to predict the correct label for the pair, although presence of a comparative or superlative is slightly more predictive of a \textsc{neutral} label. 

Moreover, the baseline models perform below average on discourse markers, such as \textit{despite} and \textit{however}, losing roughly 2 to 3 points each. Unsurprisingly, the attention-based ESIM model performs better than the other two on sentences with greater than 20 words. Additionally, our baseline models do show slight improvements in accuracy on negation, suggesting that they may be tracking it as a predictor of \textsc{contradiction}.

\section{Conclusion}

Natural language inference makes it easy to judge the degree to which neural network models for sentence understanding capture the full meanings for natural language sentences. Existing NLI datasets like SNLI have facilitated substantial advances in modeling, but have limited headroom and coverage of the full diversity of meanings expressed in English. This paper presents a new dataset that offers dramatically greater linguistic difficulty and diversity, and also serves as a benchmark for cross-genre domain adaptation. 

Our new corpus, MultiNLI, improves upon SNLI in its empirical coverage---because it includes a representative sample of text and speech from ten different genres, as opposed to just simple image captions---and its difficulty, containing a much higher percentage of sentences tagged with one or more elements from our tag set of thirteen difficult linguistic phenomena. This greater diversity is reflected in the dramatically lower baseline model performance on MultiNLI than on SNLI (see Table \ref{tab:autotag}) and comparable inter-annotator agreement, suggesting that MultiNLI has a lot of headroom remaining for future work.

The MultiNLI corpus was first released in draft form in the first half of 2017, and in the time since its initial release, work by others \citep[][]{conneau2017} has shown that NLI can also be an effective source task for pre-training and transfer learning in the context of sentence-to-vector models, with models trained on SNLI and MultiNLI substantially outperforming all prior models on a suite of established transfer learning benchmarks. We hope that this corpus will continue to serve for many years as a resource for the development and evaluation of methods for sentence understanding. 

\section*{Acknowledgments}
This work was made possible by a Google Faculty Research Award to SB and AL. SB also gratefully acknowledges gift support from Tencent Holdings. 
We also thank George Dahl, the organizers of the RepEval 2016 and RepEval 2017 workshops, Andrew Drozdov, Angeliki Lazaridou, and our other colleagues at NYU for their help and advice.

\bibliography{tacl}
\bibliographystyle{acl_natbib}

\end{document}